# Combining a Convolutional Neural Network with Autoencoders to Predict the Survival Chance of COVID-19 Patients

Fahime Khozeimeh[a], Danial Sharifrazi[b], Navid Hoseini Izadi[c], Javad Hassannataj Joloudari[d], Afshin Shoeibi[e,f], Roohallah Alizadehsani[a,*], Juan M. Gorriz[g,h], Sadiq Hussain[i], Zahra Alizadeh Sani[j], Hossein Moosaei[k], Abbas Khosravi[a], Saeid Nahavandi[a], Sheikh Mohammed Shariful Islam [l,m,n]

[a]Institute for Intelligent Systems Research and Innovation (IISRI), Deakin University, Geelong, Australia

[b]Department of Computer Engineering, School of Technical and Engineering, Shiraz Branch, Islamic Azad University, Shiraz, Iran

[c]Department of Electrical and Computer Engineering, Isfahan University of Technology, Isfahan 84156-83111, Iran

[d]Department of Computer Engineering, Faculty of Engineering, University of Birjand, Birjand, Iran

[e]Computer Engineering Department, Ferdowsi University of Mashhad, Mashhad, Iran

[f]Faculty of Electrical and Computer Engineering, Biomedical Data Acquisition Lab, K. N. Toosi University of Technology, Tehran, Iran

[g]Department of Signal Theory, Networking and Communications, Universidad de Granada

[h]Department of Psychiatry, University of Cambridge, Cambridge, UK

[i]System Administrator, Dibrugarh University, Assam 786004, India

[j]Omid hospital, Iran University of Medical Sciences, Tehran, Iran

[k]Department of Mathematics, Faculty of Science, University of Bojnord, Iran; moosaei@ub.ac.ir

[l]Institute for Physical Activity and Nutrition, School of Exercise and Nutrition Sciences, Deakin University, Geelong, VIC, 3220, Australia

[m]Cardiovascular Division, The George Institute for Global Health, Newtown, Australia

[n]Sydney Medical School, University of Sydney, Camperdown, Australia

∗Corresponding author: Roohallah Alizadehsani      Email: r.alizadehsani@deakin.edu.au

## ABSTRACT

COVID-19 has caused many deaths worldwide. The automation of the diagnosis of this virus is highly desired. Convolutional neural networks (CNNs) have shown outstanding classification performance on image datasets. To date, it appears that COVID computer-aided diagnosis systems based on CNNs and clinical information have not yet been analysed or explored. We propose a novel method, named the CNN-AE, to predict the survival chance of COVID-19 patients using a CNN trained with clinical information. Notably, the required resources to prepare CT images are expensive and limited compared to those required to collect clinical data, such as blood pressure, liver disease, etc. We evaluated our method using a publicly available clinical dataset that we collected. The dataset properties were carefully analysed to extract important features and compute the correlations of features. A data augmentation procedure based on autoencoders (AEs) was proposed to balance the dataset. The experimental results revealed that the average accuracy of the CNN-AE (96.05%) was higher than that of the CNN (92.49%). To demonstrate the generality of our augmentation method, we trained some existing mortality risk prediction methods on our dataset (with and without data augmentation) and compared their performances. We also evaluated our method using another dataset for further generality verification. To show that clinical data can be used for COVID-19 survival chance prediction, the CNN-AE was compared with multiple pre-trained deep models that were tuned based on CT images.

Keywords: COVID-19, survival chance, CNN, Autoencoder, data augmentation, feature analysis

# 1 Introduction

Currently, medical centres hold huge amounts of patient data. Medical biomarkers, demographic data and image modalities can help and support medical specialists to diagnose infectious diseases [1], Alzheimer's [2], Parkinson [3] and coronary artery disease [4]. However, these data must be processed and analysed if they are to become usable information for specialists. Automated solutions based on artificial intelligence have the potential to carry out the required process efficiently [5].

Recently, a new type of coronavirus (i.e., Coronavirus Disease 2019 [COVID-19]) emerged, which has taken many lives worldwide [6-9]. The virus outbreak was observed for the first time in late 2019 [10, 11]. COVID-19 primarily targets the lungs [12, 13]. Thus, if the virus is not properly diagnosed in the early stages of infection, it can severely damage the lungs [14]. The mortality rate of the virus is low; however, it must not be overlooked, as the virus is highly contagious. The virus threat becomes more serious when the resources of medical centres cannot provide services to the large number of people who are infected each day [15].

The prediction of the survival chance of infected individuals is as important as the early detection of the virus. Under resource scarcity, medical centres can take into account patients' conditions and use the available resources wisely. Previous research on COVID-19 detection has proven that deep neural networks are very effective in the early detection of COVID-19 [16]. Thus, it may be that deep networks are also useful for survival chance prediction. In this study, we relied on a clinical dataset, which included data about gender, age and blood type, to perform a diagnostic analysis of the COVID-19 virus. To the best of our knowledge, this appears to be the first paper to propose a survival chance predictor for COVID-19 patients using clinical features. To evaluate the effectiveness of our proposed method, we compared its performance against a standard convolutional neural network (CNN) trained on image data. This study makes a number of contributions as follows:

- The survival chance prediction of COVID-19 patients based on clinical features
- Preparing clinical dataset to predict the survival chance of COVID-19 patients for the first time
- Providing a careful analysis of the dataset characteristics, including an examination of the effects of features on the mortality rate and the correlations between each feature pair
- Making our dataset publicly available
- Combining Autoencoder (AE) with CNN to increase prediction accuracy
- Proposing a data augmentation procedure to balance the number of samples of different classes of the dataset. Notably, our data augmentation method is generic and applicable to any other dataset.

The remaining sections of the paper are organised as follows: Section 2 reviews the related literature; Section 3 briefly sets out the required background; Section 4 describes our dataset; Section 5 explains the proposed methodology; Section 6 presents our experimental results; and Sections 7 and 8 present our discussion, conclusion and future works.

# 2 Literature Review

This study sought to predict the survival chance of COVID-19 patients using clinical features. We began by reviewing the COVID-19 detection methods that rely on clinical features and image data. We also reviewed methods on mortality estimations of infected patients.

To contain the COVID-19 threat as soon as possible, researchers approached this virus from multiple directions. Some focused on the fast and accurate detection of infected patients. For example, Wu et al.

[17] extracted 11 vital blood indices using the random forest (RF) method to design an assistant discrimination tool. Their method had an accuracy of 96.97% and 97.95% for the test set and cross-validation set, respectively. The assistant tool was well equipped to perform a preliminary investigation of suspected patients and suggest quarantine and timely treatment. In another study, Rahman et al. [18] reviewed various studies on treatment, complications, seasonality, symptoms, clinical features and the epidemiology of COVID-19 infection to assist medical practitioners by providing necessary guidance for the pandemic. Using a CNN, they tried to detect infected patients to isolate them from healthy patients.

Various hybrid approaches have been adopted to improve COVID-19 diagnosis accuracy. Islam et al. [19] employed a CNN for feature extraction and long short-term memory for the classification of patients based on X-ray images. EMCNet [20] is another hybrid diagnosis approach that uses a CNN for feature extraction and carries out binary classification using a number of learning techniques, including RF and support vector machine (SVM), on X-ray images. Islam et al. [21] also used a CNN for feature extraction but relied on a recurrent neural network (RNN) for classification based on the extracted features. Multiple experiments have been conducted using a combination of architectures, such as VGG19 and DenseNet121, with an RNN. VGG19+RNN was reported to have the best performance.

In addition to distinguishing between infected and non-infected patients, it is also important to determine whether infected patients have severe conditions. Muhammad et al. [22] relied on data mining to predict the recovery condition of infected patients. Their method was able to determine the age group of high-risk patients who are less likely to recover and those who are likely to recover quickly. Their method was able to provide the minimum and the maximum number of days required for a patient's recovery. Chen et al. [23] studied 148 severe and 214 non-severe COVID-19 patients from Wuhan, China using their laboratory test results and symptoms as features to design a RF. The task of the RF was to classify COVID-19 patients into severe and non-severe types using the features. Using the laboratory results and symptom as input, the accuracy of their model was over 90%. Some of the key features they identified were lactate dehydrogenase (LDG), interleukin-6, absolute neutrophil count, D-Dimer, diabetes, gender, cardiovascular disease, hypertension and age.

Other researchers have focused on the mortality risk prediction of the patients. Gao et al. [24] proposed a mortality risk prediction model for COVID-19 (MRPMC) that applied clinical data to stratify patients by mortality risk and predicted mortality 20 days in advance. Their ensemble framework was based on four machine-learning techniques; that is, a neural network (NN), a gradient-boosted decision tree [25], a SVM and logistic regression. Their model was able to accurately and expeditiously stratify the mortality risk of COVID-19 patients.

Zhu et al. [26] presented a risk stratification score system as a multilayer perceptron (MLP) with six dense layers to predict mortality. 78 clinical variables were identified and prediction performance was compared with the pneumonia severity index, the confusion, uraemia, respiratory rate, BP, age ≥ 65 years score and the COVID-19 severity score. They derived the top five predictors of mortality; that is, LDH, C-reactive protein, the neutrophil to lymphocyte ratio, the Oxygenation Index and D-dimer. Their model was proved to be effective in resource-constrained and time-sensitive environments.

The power of the XGBoost algorithm has also been leveraged for mortality risk prediction. For example, Yan et al. [27] collected blood samples of 485 infected patients from China to detect key predictive biomarkers of COVID-19 mortality. They employed a XGBoost classifier that was able to predict the mortality of patients with 90% accuracy more than 10 days in advance. In another study, Bertsimas et al. [28] developed a data-driven mortality risk calculator for in-hospital patients. Laboratory, clinical and demographic variables were accumulated at the time of hospital admission. Again, they applied XGBoost to predict the mortality of patients. Adopting a different approach, Abdulaal et al. [29] devised a point-of-admission mortality risk scoring system using a MLP for COVID-19 patients. The network exploited patient specific features, including present symptoms, smoking history, comorbidities and demographics, and predicted the mortality risk based on these features. The mortality prediction model demonstrated a specificity of 85.94%, a sensitivity of 87.50% and an accuracy of 86.25%.

As the symptoms of different viruses may be similar to some extent, there has been an attempt to distinguish different viruses from one another [30]. To this end, multiple classical machine-learning algorithms were trained to classify textual clinical reports into the four classes of Severe acute respiratory syndrome (SARS), acute respiratory distress syndrome, COVID-19 and both SARS and COVID-19. Feature engineering has also been carried out using report length, bag of words and etc. Multinomial Naïve Bayes and logistic regression outperformed other classifiers with a testing accuracy of 96.2%. A summary of the reviewed works are presented in Table 1.

Most existing studies on COVID-19 have relied on computed tomography (CT) and X-ray images to achieve their research objectives. Al-Waisy et al. [31] proposed COVID-DeepNet, a hybrid multimodal deep-learning system for diagnosing COVID-19 using chest X-ray images. After the pre-processing phase, the predictions from two models (a deep-belief network and a convolutional deep-belief network) were fused to improve diagnosis accuracy. Another fusion of two models (ResNet34 and a high-resolution network model) was proposed in [32] to form the COVID-CheXNet method for COVID-19 diagnosis. Mohammed et al. collected a dataset of X-ray images and made it publicly available. The dataset has been used to benchmark various machine-learning methods for COVID-19 diagnosis [33]. They reported that the ResNet50 model achieved the best performance. In another benchmarking study [34], 12 COVID-19 diagnostic methods were examined based on 10 evaluation criteria. To this end, multicriteria decision making (MCDM) and the technique order of preference by similarity to ideal solution were employed. The 10 criteria were weighted based on entropy. The SVM classifier was reported to have the best performance among the benchmarked methods.

Slowing down the spread of COVID-19 and supporting infected patients are as important as COVID-19 detection. Several works have investigated the possibility of using existing technologies to benefit infected patients. Rahman et al. [35] proposed a deep-learning architecture to determine whether people are wearing a facial mask. The monitoring was realised via closed-circuit television cameras in public places. Islam et al. [36] reviewed existing technologies that can facilitate the breathing of infected patients. Wearable technologies and how they can be used to provide initial treatment to people have also been investigated [37]. Ullah et al. [38] reviewed telehealth services and the possible ways in which they can be used to provide patients with necessary treatments while keeping the social distance between patients and doctors.

Some works have adopted a broader approach and reviewed various recently developed deep-learning methods with application to COVID-19 diagnosis. For example, Islam et al. [39] reviewed these methods based on X-ray and CT images while the overall application of deep learning for diagnosis purposes to control the pandemic threat has been discussed in [40].

Based on the review presented above, it is apparent that existing works based on clinical data are rather scarce. Thus, we sought to conduct another study using clinical data for mortality risk assessment. The difference between our method and existing research on mortality risk assessment is twofold. First, we developed a new approach for carrying out the assessment. Second, some of the clinical features that we considered had never been used previously, which is why we have released our dataset publicly. As will be discussed further below, clinical data are more cost effective than CT images, and classifiers trained on clinical data achieve a level of performance that is almost equal to that achieved by classifiers trained on CT images. To justify this claim, we compared the performance of our method trained on clinical data to a standard CNN trained on CT images.

Table 1. Summary of the reviewed literature

| Ref | Method | objective |
| --- | --- | --- |
| Gao et al. [24] | An Ensemble of NN, grad boosted decision tree, SVM, and logistic regression | mortality risk prediction |
| Zhu et al. [26] | MLP | mortality risk prediction |
| Yan et al. [27] | XGBoost classifier | mortality risk prediction |
| Bertsimas et al. [28] | XGBoost classifier | mortality risk prediction |
| Abdulaal et al. [29] | MLP | mortality risk prediction |
| Wu et al. [17] | RF | COVID-19 detection |
| Rahman et al. [18] | CNN | COVID-19 detection |
| Khanday et al. [30] | Multinomial Naïve Bayes and Logistic regression | Patients classification into four classes {SARS, ARDS, COVID-19, Both (SARS, COVID-19)} |
| Chen et al. [23] | RF | COVID-19 severity classification |

## 3 Background

Our proposed method comprises two modules: the classifier and data augmenter. The classification is carried out using a CNN. The data augmentation is realised using 10 AEs. In this section, we briefly review the main concepts of CNNs and AEs.

### 3.1 CNNs

CNNs are massively used in image-based learning applications. Due to the automatic feature extraction mechanism of CNNs, they can discover valuable information from training samples. CNNs are usually designed with several convolutional, pooling and fully connected layers [41]. As Figure 1 shows, feature extraction is done by convolving the input with convolutional kernels. The pooling layer reduces the computational volume of the network without making a noticeable change in the resolution of the feature map. In CNNs, the size of the pooling layers usually decreases as the number of layers increases. Two of the most popular types of pooling layers are max pooling and average pooling [42].

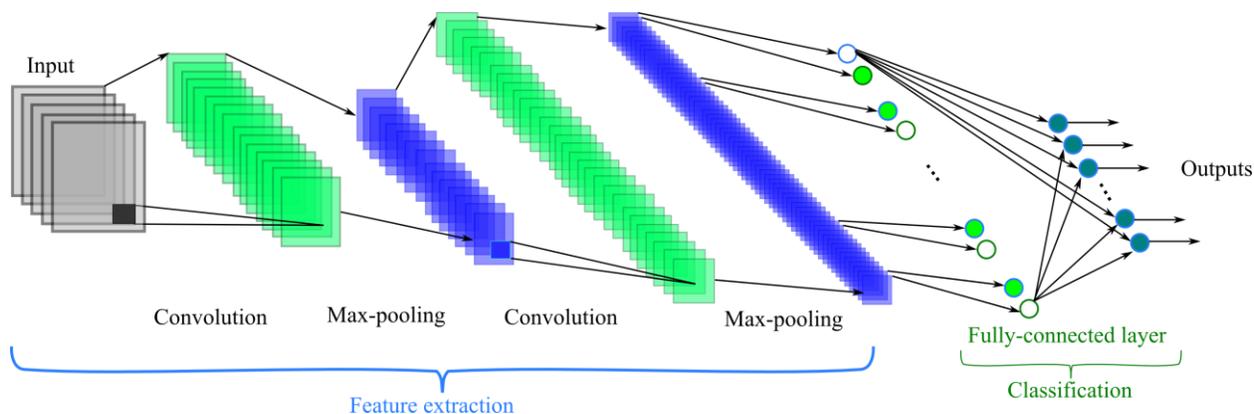

Figure 1. A CNN schematic.

## 3.2 AEs

AEs belong to the realm of unsupervised learning, as they do not need labelled data for their training. In brief, an AE compresses input data to a lower dimensional latent space and then reconstructs the data by decompressing the latent space representation. Similar to principle component analysis (PCA), AEs perform dimensionality reduction in the compression phase. However, unlike PCA, which relies on linear transformation, AEs carry out nonlinear transformation using deep neural networks [43]. Figure 2 shows the architecture of a typical AE.

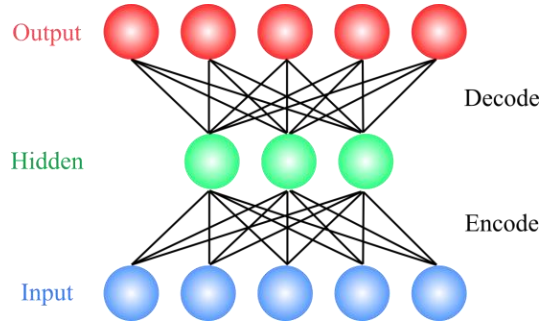

Figure 2. AE architecture: High-dimension input data are encoded (compressed) to form a latent (hidden) space that has a lower dimension than that of the original input. The latent representation is reconstructed (decoded) to yield decompressed output.

## 3.3 Information Gain

In this section, we review information gain (IG), as it is used to determine the degree to which each feature of our dataset contributes to the patients' deaths (see Section 4). IG calculates the entropy reduction that results from splitting a dataset, $D$, based on a given value, $a$, of a random variable, $A$, such that:

$$IG(D, A = a) = H(D) - H(D|A = a),$$

where $H(D)$ and $H(D|A = a)$ are entropy on dataset $D$ and conditional entropy on dataset $D$, respectively, given that $A = a$.

Conditional entropy is computed as:

$$H(D|A = a) = \sum_{v \in values(A)} \frac{|D_{A=a}|}{|D|} H(D_{A=a}), \qquad (1)$$

where $D_{A=a} \subset D$ is the set of samples with variable $A = a$ and $|D_{A=a}|$ and $|D|$ denote the cardinality of subset $D_{A=a}$ and set $D$, respectively. In Equation (1), the sum is computed over all possible values of $A$.

## 4 Description of our clinical dataset

The dataset we collected in this paper comprised 320 patients (300 cases of recovered patients and 20 cases of deceased patients). The percentage of female cases was 55%. The mean age of patients in the dataset was 49.5 years old, and the standard deviation was 18.5. The patients referred to Tehran Omid hospital in Iran from 3 March 2020 to 21 April 2020. Ethical approval for the use of these data was obtained from the Tehran Omid hospital. In gathering the data, patients' history (as collected by doctors), questionnaires (as completed by patients), laboratory tests, and vital sign measurements were used. Descriptions of the dataset features are presented in Table 2. Our dataset is publicly available in [44]. Institutional approval was granted for the use of the patient datasets in research studies for diagnostic and therapeutic purposes. Approval was granted on the grounds of existing datasets. Informed consent was obtained from all of the patients in this study. All methods were carried out in accordance with relevant guidelines and regulations.

Table 2. Description of the dataset features used for classification.

| Feature Name | Range |
|---|---|
| Gender | {Male, Female} |
| Age | 11-95 years old |
| Blood Type | {A-, A+, B-, B+, AB-, AB+, O-, O+} |
| BCG Vaccine | {Yes, No} |
| CBC | {Normal, Abnormal} |
| Diabetes | {Yes, No} |
| blood pressure | {Yes, No} |
| Asthma | {Yes, No} |
| Heart disease | {Yes, No} |
| kidney disease | {Yes, No} |
| Respiratory disease | {Yes, No} |
| Cancer | {Yes, No} |
| Corticosteroids | {Yes, No} |
| Transplant | {Yes, No} |
| HEM | {Yes, No} |
| Immunodeficiency | {Yes, No} |
| Liver disease | {Yes, No} |
| Rheumatological disease | {Yes, No} |
| Chest pain | {Yes, No} |
| Fever | {Yes, No} |
| Trembling or Shakes | {Yes, No} |
| Weakness | {Yes, No} |
| Sweating | {Yes, No} |
| Sore throat | {Yes, No} |
| Dyspnea | {Yes, No} |
| Dry cough | {Yes, No} |
| Cough with sputum | {Yes, No} |
| Fatigue, whole body hurts | {Yes, No} |
| Anosmia | {Yes, No} |
| Ageusia | {Yes, No} |
| Anorexia | {Yes, No} |
| Eczema | {Yes, No} |
| Conjunctivitis (Pink eye) | {Yes, No} |
| Blindness and Tunnel vision | {Yes, No} |
| Vertigo | {Yes, No} |
| Nausea/Diarrhea | {Yes, No} |
| Tobacco | {Yes, No} |

As our dataset had not been released previously, it was vital to assess the degree to which each dataset feature contributed to patients' deaths. Such an analysis provides researchers with valuable insights into the characteristics of the collected data. Various feature selection methods are available to determine the weight of each feature in the classification of dataset samples. We chose IG [45], which is one of the most widely used feature selection methods [46]. In Figure 3, the importance of each feature (i.e., the IG) is shown as a bar. Age had a much larger IG (0.149) than other features. Thus, age was not included in Figure 3 to make it easier to compare the importance of the other features. According to the bar chart, (after age) cancer, heart and kidney diseases were the second, third and fourth most important features related to patients' deaths, respectively. Thus, it was clear that patients with poor health conditions were more vulnerable to COVID-19. It should be noted that Figure 3 does not include the features with zero IG.

We also inspected the interplay between the dataset features to determine the potential correlation between them. To this end, the grid in Figure 4 is presented. Figure 4 can be thought as a heat map that shows the positive/negative correlation between features. Each cell $c(i,j)$ in the grid of Figure 4 represents the correlation of features in the i-th row and j-th column. As the cell colour approaches red, the positive correlation between the feature pairs is higher. For example, anosmia (the loss of the ability to smell) and ageusia (the loss of the ability to taste with the tongue) had a high positive correlation, which means they were usually observed simultaneously.

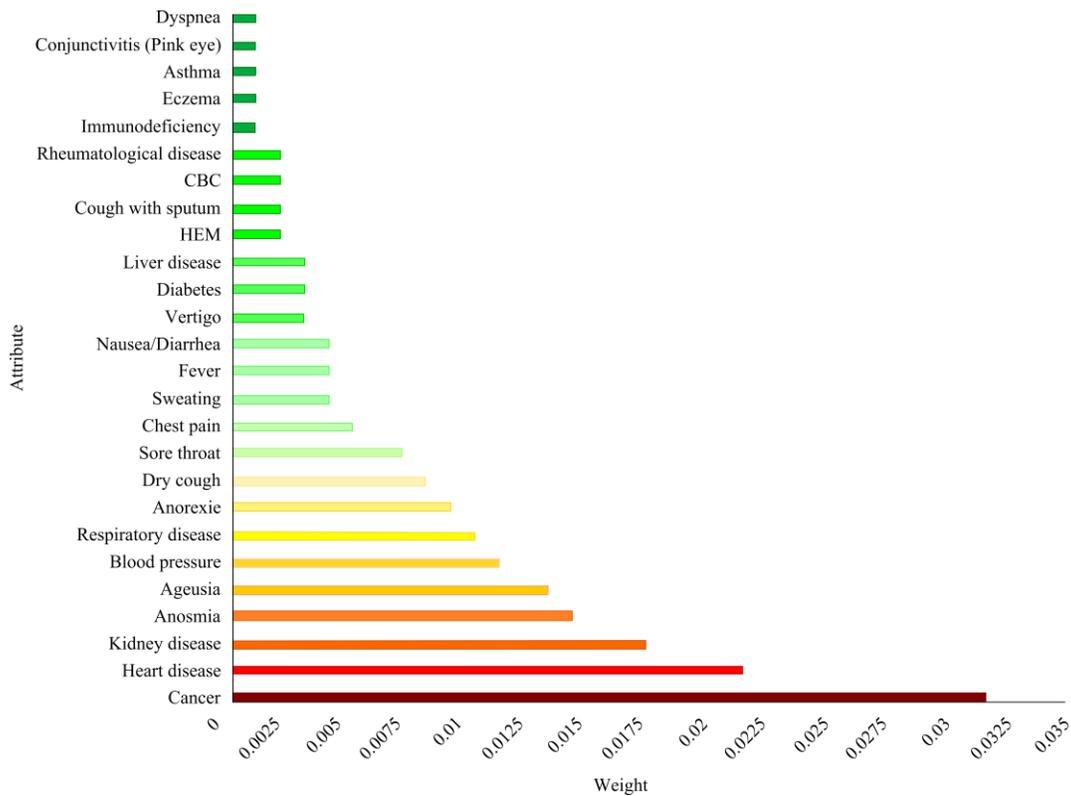

Figure 3. Feature effects on mortality rate based on IG.

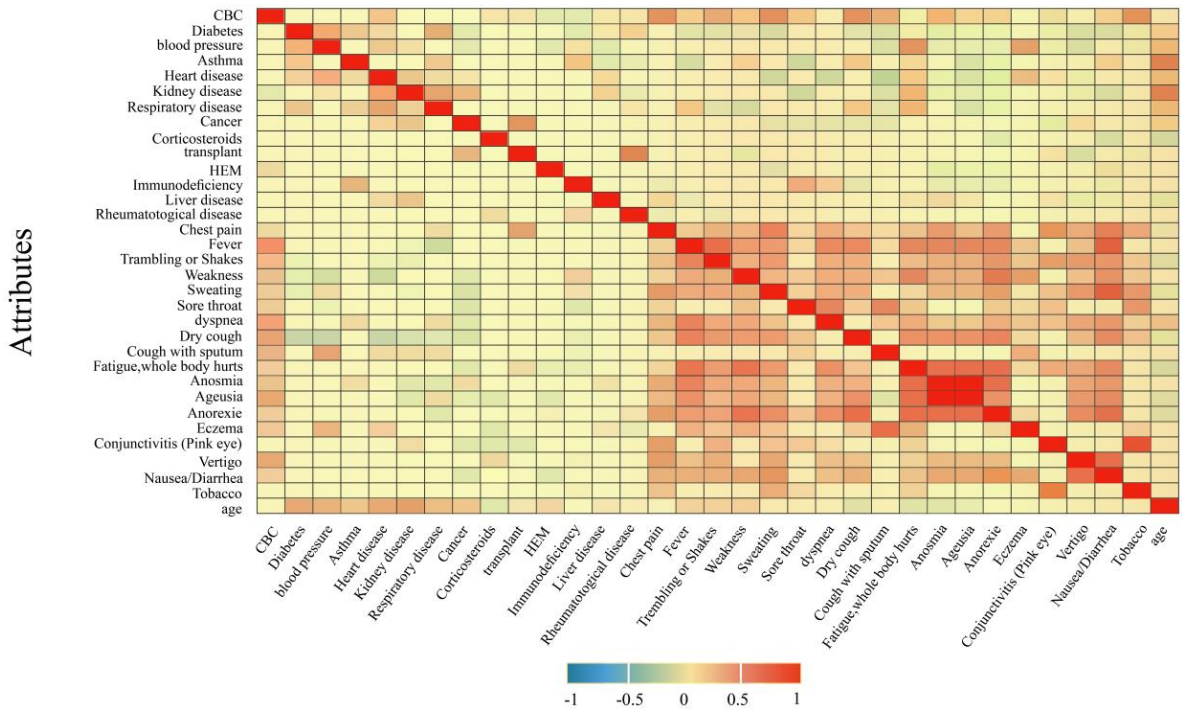

Figure 4. Correlation between dataset features.

## 5  Proposed Methodology

This study investigated the survival chance prediction of COVID-19 patients who referred to the Omid hospital in Tehran. The classification was based on features obtained from patients' information. In the dataset collected, the number of recovered patients was 300 and the number of deceased patients was 20. The number of recovered patients was clearly much higher than that of the deceased patients. To ensure accurate classification, it was necessary to balance the recovered to the deceased ratio of the dataset samples. To do this, the number of instances of the lower class was increased, such that the number of data in both classes was approximately equal. To increase the number of data of deceased patients, an AE model was used. To carry out the data augmentation, the 20 samples of the deceased class were fed to the AE to undergo the compression and decompression routines. The output of this process comprised 20 reconstructed samples that were similar (but not identical) to the original ones. Thus, we augmented the original 20 samples with 20 reconstructed samples. Training the AE 10 times using different training and validation sets yielded 10 AEs with a similar architecture but different parameters. Each of the 10 AEs generated 20 reconstructed deceased samples, yielding reconstructed samples of 200 overall, which were added to the original deceased samples. To provide an insight into the function of the AEs, sample vectors before and after reconstruction are presented in Table 3. For the majority of '1' elements of input vector $c$, the AE outputted values near 1 as the elements of reconstructed vector $\hat{c}$. Similarly, most of the reconstructed elements corresponding to original '0' elements had values near '0', which shows that the reconstruction process was sound.

Table 3. An example of reconstruction performed by an AE: Vector $c$ is the original sample and vector $\hat{c}$ is its reconstructed counterpart.

| $c[1:10]$ | 1 | 0 | 0 | 1 | 0 | 0 | 0 | 0 | 0 | 0 |
|---|---|---|---|---|---|---|---|---|---|---|
| $\hat{c}[1:10]$ | 0.9940 | 0.1291 | 0.0001 | 0.4697 | 0.1581 | 0.0240 | 0.0525 | 0.0068 | 0.0061 | 0.0202 |
| $c[11:20]$ | 0 | 0 | 0 | 1 | 0 | 0 | 0 | 0 | 1 | 1 |
| $\hat{c}[11:20]$ | 0 | 0 | 0.0003 | 0.4004 | 0.0004 | 0.0596 | 0.0040 | 0.0027 | 0.9516 | 0.4450 |
| $c[21:30]$ | 0 | 0 | 0 | 1 | 1 | 0 | 1 | 0 | 0 | 0 |
| $\hat{c}[21:30]$ | 0.1305 | 0.0018 | 0.0042 | 0.9565 | 0.5750 | 0.0029 | 0.9281 | 0.0111 | 0.0140 | 0.0966 |
| $c[31:39]$ | 0 | 0 | 0 | 0 | 0 | 0 | 0 | 0 | 1 | - |
| $\hat{c}[31:39]$ | 0.0087 | 0.0004 | 0 | 0.0110 | 0.0024 | 0 | 0.0017 | 0.0015 | 0.9814 | - |

The details of the augmentation process are explained in more detail in Subsection 5.1. It should be noted that our augmentation procedure is generic and can be applied to any other dataset.

## 5.1 Implementation details of CNN-AE

The proposed CNN-AE method comprises multiple steps (see Figure 5 for a summary). The pseudo-code of the method is also available in Algorithm 1. The detailed explanation of the pseudo-code is presented below:

1. 10 AEs $\{AE_1, \ldots, AE_{10}\}$ were designed with identical configuration but different initial parameters for data augmentation (line 1).
2. Each of the 10 AEs was trained on 300 samples representing the recovered patients. Our objective was to have 10 models with different parameters at the end of the training. To this end, we divided the 300 samples into 10 groups of 30 samples $\{g_j, j = 1, 2, \ldots, 10\}$ where $g_j$ is the j-th group of samples. To train the i-th model, $g_i$ was set aside for validation and the nine remaining groups $\{g_j, j \in \{1, 2, \ldots, 10\} - \{i\}\}$ (270 samples) were used for training. It should be noted that each model was initialised with different parameters, trained on partially different training samples and validated on a totally different validation set. Thus, the proposed training procedure yielded 10 different AEs (lines 2–4).
3. The 20 deceased samples were fed to each of the 10 trained AEs. The samples underwent the compression and decompression routine of the AEs. As the decompression procedure was lossy, the 20 reconstructed samples (after decompression) were not identical to the original samples. Additionally, the 10 trained AEs exhibited different behaviours on the same input data, as their parameters were different from each other. Thus, feeding the same 20 samples to the 10 AEs yielded 200 new samples that belonged to the deceased class (lines 5–8). The explained procedure sought to augment the data to remedy the lack of sufficient samples for the deceased class.
4. The 200 reconstructed samples were attached to 320 original samples to yield a dataset of 520 samples (line 9).
5. A CNN model was designed to classify 520 samples as recovered or deceased (line 10).
6. The CNN model was trained using all 520 samples. A 10-fold cross-validation was applied during the training (lines 11–20). Thus, the training sample size was 468 (samples of 9 folds), and the test sample size was 52 (samples of 1-fold).
7. The trained CNN was used to classify the test data (line 21).

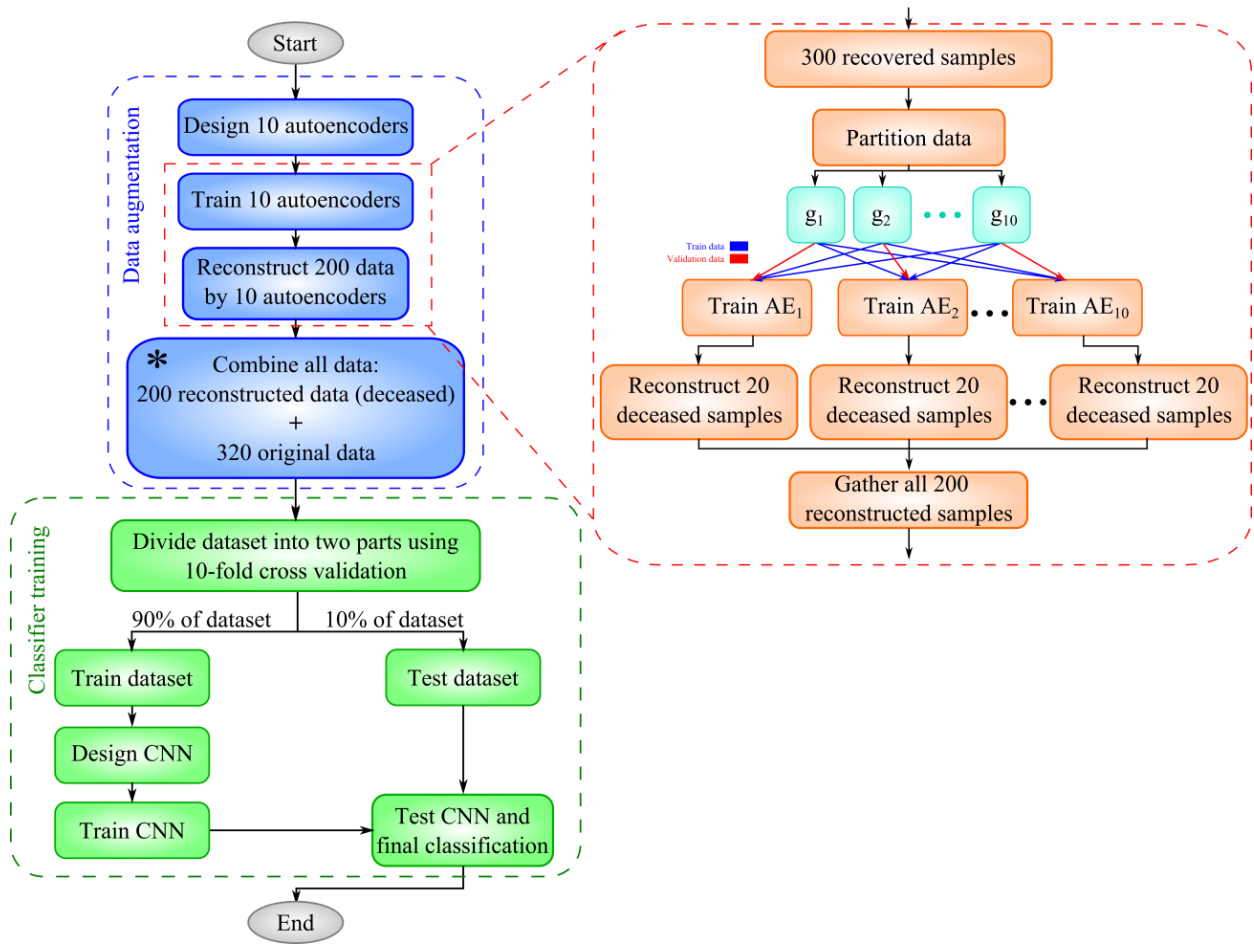

Figure 5. The steps for implementing the proposed method.

| Algorithm 1. CNN-AE pseudo-code |  |
|---|---|
| **Input:** dataset $D = \{D_{recovered} \cup D_{deceased}\}$, training epochs N, batch size B, number of folds K | |
| | `// Auto-encoders initialization` |
| 1: | Create 10 autoencoders with initial random parameters: $\{AE_1, \dots, AE_{10}\}$ |
| | `// Autoencoders training` |
| 2: | Partition samples in $D_{recovered}$ to 10 subsets: $\{g_1, \dots, g_{10}\}$ |
| 3: | For i=1:10 |
| 4: | Train $AE_i$ on $D_{recovered} - g_i$ and perform validation on $g_i$ |
| | `// Augmented data generation` |
| 5: | $A = []$ |
| 6: | For i=1:10 |
| 7: | $a_i = AE_i(D_{deceased})$ |
| 8: | $A = A \cup a_i$ |
| 9: | $D_{augmented} = D \cup A$ |
| 10: | Create CNN $C$ with initial random parameters |
| | `// K-Fold cross validation` |
| 11: | Partition $D_{augmented}$ to 90% training set $D_{train}$ and 10% test set $D_{test}$ |
| 12: | Partition $D_{train}$ to K subsets $\{F_1, \dots, F_K\}$ |
| 13: | For k=1:K |
| 14: | $\quad D_{train} = D_{augmented} - F_K$ |
| 15: | $\quad D_{valid} = F_K$ |
| 16: | $\quad$ For e=1:N |
| 17: | $\quad\quad batch_t = $ sample_batch$(D_{augmented}, B)$ |
| 18: | $\quad\quad$ CNN.train$(batch_t)$ |
| 19: | $\quad\quad batch_v = $ sample_batch$(D_{valid}, B)$ |
| 20: | $\quad\quad$ CNN.validate$(batch_v)$ |
| 21: | CNN.test$(D_{test})$ |
| 22: | **Return** CNN |

To implement the proposed method, we used Python language and the Keras library, which has a TensorFlow backend. In this study, the dataset contained 320 samples of infected cases. Of these 320 cases, the number of recovered cases was 300, and the number of deceased cases was 20. Additionally, we also generated 200 reconstructed deceased cases to balance the recovered to the deceased ratio of our dataset. After the reconstruction phase, our dataset contained 520 cases. We used a 10-fold cross-validation. Additionally, 80% of 9 of the folds were used for training, and the remaining 20% was used for validation. The implementation details of CNN and AE are illustrated in Figures 6 and 7, respectively.

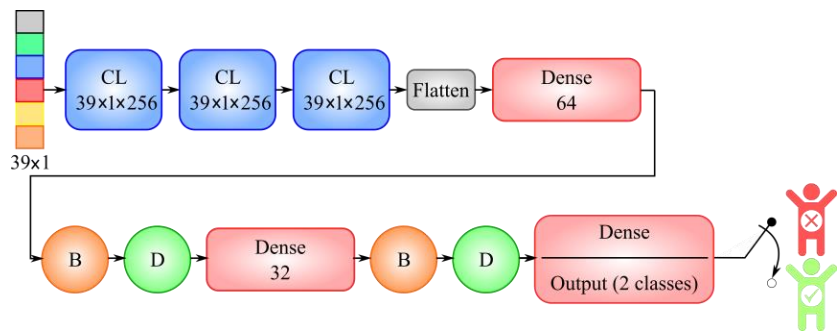

Figure 6. Implementation details of CNN. 'CL' and 'Dense' represent convolutional and fully connected layers, respectively. Circles with the letter 'B' represent batch normalisation layers, and circles with the letter 'D' represent dropout layers with a probability 0.5.

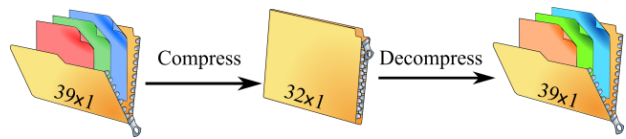

Figure 7. The implemented AE: The 39-dimensional input vector was compressed to a 32-dimensional vector. During reconstruction, the 32-dimensional vector was decompressed to a 39-dimensional vector.

# 6 Experiments

In this section, the experimental results are presented. The implementation details of CNN and AEs are explained in Section 6.1. We report on the performance of the proposed method (CNN-AE) and compare it to a CNN in Section 6.2.

## 6.1 Experimental details

Our experiments consisted of two scenarios. In the first scenario, our CNN-AE method was compared to a standard CNN method that was trained on clinical data. The architecture of the CNN is presented in

Table 4. To ensure a fair comparison, we used the same CNN architecture in our method. The implementation details of the AEs used in the CNN-AE are presented in Table 5.

Table 4. Implementation details of the CNN trained on clinical data.

| Hyper-parameters | Values |
|---|---|
| Input dimension | $39 \times 1$ (39 medical features) |
| Number of convolutional layers | 3 |
| Number of fully connected layers | 3 |
| Number of filters for each convolutional layer | 256 |
| Size of convolutional kernels | $3 \times 3$ |
| Strides size | 1 |
| Activation function for hidden layers | ReLU |
| Activation function of the last layer | Sigmoid |
| Adam hyper-parameters | $\beta_1 = 0.9, \beta_2 = 0.999$ |
| Learning rate | 0.001 |
| Loss function | Binary Cross Entropy (BCE) |
| Number of neurons of fully connected layers | 64, 32, 2 |
| Dropout probability | 0.5 |
| Number of epochs | 100 |

Table 5. AE implementation details.

| Hyper-parameters | values |
|---|---|
| Input dimension | $39 \times 1$ (39 medical features) |
| Number of neurons of the first layer | $39 \times 1$ |
| Number of neurons of the second layer | $32 \times 1$ |
| Number of neurons of the third layer | $39 \times 1$ |
| First and second layers activation function | ReLU |
| Third layer activation function | Sigmoid |
| Adam hyper-parameters | $\beta_1 = 0.9, \beta_2 = 0.999$ |
| Learning rate | 0.001 |
| Loss function | Binary Cross Entropy (BCE) |
| Number of epochs | 100 |

In the second phase of our experiments, we compared the CNN-AE trained on clinical data to a standard CNN trained on image data. The CNN architecture is presented in Figure 8. After multiple trials, we obtained the best set of the CNN hyperparameters (see Table 6).

Table 6. Implementation details of the CNN trained on image data.

| Hyper-parameters | Values |
| --- | --- |
| Number of convolutional kernels of first layer | 64 |
| Number of convolutional kernels of second layer | 128 |
| Number of convolutional kernels of third layer | 256 |
| Size of convolutional kernels | $3 \times 3$ |
| Strides size | 2 |
| Input dimension | $100 \times 100$ |
| Output dimension | 2 |
| Number of convolutional layers | 3 |
| Number of fully connected layers | 2 |
| Activation function for convolutional and fully connected layers | ReLU |
| Activation function of last layer | Sigmoid |
| Adam hyper-parameters | $\beta_1 = 0.9, \beta_2 = 0.999$ |
| Learning rate | 0.001 |
| Loss function | Binary Cross Entropy (BCE) |
| Number of neurons of the fourth layer (fully connected) | 256 |
| Number of neurons of fifth layer (fully connected) | 128 |
| Dropout probability | 0.5 |
| Number of epochs | 30 |
| Batch size | 128 |

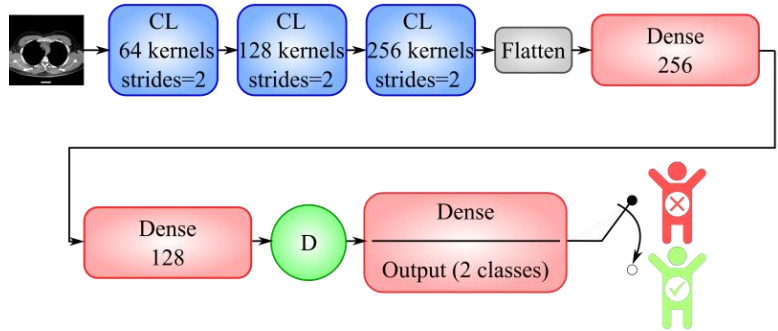

Figure 8. Implementation details of the CNN trained on CT images. 'CL' and 'Dense' represent convolutional and fully connected layers, respectively. Circles with the letter 'D' represent dropout layers with a probability 0.5.

## 6.2 Experimental results

We sought to answer two important questions about the proposed method. First, we compared our method performance with a standard CNN trained on clinical data. This experiment examined the effects of the proposed data augmentation technique using multiple AEs. We also trained a standard CNN for the same purpose (to predict patients' survival chance) but used CT images. This experiment sought to determine how well CT images can represent patients' survival chance using a CNN as the predictor.

### 6.2.1 Examining the data augmentation approach

As mentioned in Section 5.1, we used 10 AEs to augment the available dataset. Data augmentation is critical to successful training when the number of samples from different classes is unbalanced. Data imbalance can defeat any powerful classifier even a state-of-the-art CNN, which is why we employed the data augmentation technique.

To investigate the effectiveness of our data augmentation procedure, we trained a CNN on the original dataset and our CNN-AE on an augmented dataset. The original dataset comprised only 20 samples with the deceased label, but had 300 samples with the recovered label. Comparing the 300 to 20 reveals severe data imbalance from which the CNN suffered during training (see Table 7). However, using an augmented dataset with 300 recovered samples and 220 deceased samples facilitated the CNN training and improved accuracy (see Table 7). Additionally, the area under the curve (AUC) measure of the CNN-AE was almost twice that of the CNN. The specificity measure of CNN was almost zero, which was due to the fact that the CNN was unable to distinguish between deceased and recovered samples due to the insufficient number of deceased samples in the original dataset. As Table 7 shows, the CNN-AE training took more time; however, this was due to the time it took to train the 10 AEs required for data augmentation.

Table 7. Comparison of the CNN and the CNN-AE using different evaluation metrics based on a 10-fold cross-validation.

| Methods | Fold No. | Accuracy (%) | PPV (%) | Recall (%) | Specificity (%) | F1-score (%) | AUC (%) | Loss | Total training time (s) |
|---|---|---|---|---|---|---|---|---|---|
| CNN | 1 | 93.75 | 94 | 100 | 0 | 97 | 50 | 0.2157 | 27.04 |
|  | 2 | 93.75 | 94 | 100 | 0 | 97 | 50 | 0.2104 | 20.48 |
|  | 3 | 93.75 | 94 | 100 | 0 | 97 | 50 | 0.2733 | 21.34 |
|  | 4 | 90.63 | 97 | 94 | 0 | 95 | 46.77 | 0.6140 | 21.49 |
|  | 5 | 96.88 | 97 | 100 | 66.67 | 98 | 83.33 | 0.0916 | 21.70 |
|  | 6 | 90.63 | 97 | 94 | 0 | 95 | 46.77 | 0.2164 | 22.00 |
|  | 7 | 93.75 | 94 | 100 | 0 | 97 | 50 | 0.1642 | 21.51 |
|  | 8 | 93.75 | 94 | 100 | 0 | 97 | 50 | 0.1816 | 21.83 |
|  | 9 | 81.25 | 96 | 81 | 80 | 88 | 80.74 | 0.7327 | 21.97 |
|  | 10 | 96.77 | 97 | 100 | 0 | 98 | 50 | 0.1214 | 22.75 |
| 95% confidence interval over 10 folds | | 92.49±2.75 | 95.40±0.88 | 96.90±3.73 | 14.67±19.00 | 95.90±1.82 | 55.76±8.54 | 0.282±0.13 | 22.21±0.37 |
| CNN-AE | 1 | 98.08 | 97 | 100 | 94.74 | 99 | 97.37 | 0.0925 | 33.01 |
|  | 2 | 94.23 | 94 | 97 | 90.91 | 95 | 93.79 | 0.2600 | 31.50 |
|  | 3 | 100 | 100 | 100 | 100 | 100 | 100 | 0.0096 | 31.35 |
|  | 4 | 96.15 | 96 | 96 | 95.83 | 96 | 96.13 | 0.2600 | 31.60 |
|  | 5 | 93.27 | 91 | 97 | 85 | 94 | 90.94 | 0.4017 | 31.90 |
|  | 6 | 92.31 | 94 | 94 | 90.48 | 94 | 92.01 | 0.3678 | 31.99 |
|  | 7 | 98.08 | 97 | 100 | 95.45 | 98 | 97.73 | 0.0858 | 32.30 |
|  | 8 | 96.15 | 94 | 100 | 90.91 | 97 | 95.45 | 0.2027 | 32.90 |
|  | 9 | 94.23 | 92 | 96 | 92.59 | 94 | 94.3 | 0.1572 | 33.13 |
|  | 10 | 98.04 | 97 | 100 | 95.45 | 98 | 97.73 | 0.0614 | 33.83 |
| 95% confidence interval over 10 folds | | 96.05±1.48 | 95.2±1.63 | 98±1.33 | 93.14±2.52 | 96.5±1.27 | 95.55±1.70 | 0.19±0.07 | 32.35±0.49 |

In Table 7, the CNN-AE method had an average accuracy of **96.05%** and thus outperformed the CNN method, which had an average accuracy of **92.49%**. Additionally, due to the augmented data, our method was able to reduce the training/validation loss faster than CNN (as is evident in Figure 9). Similarly, the CNN-AE reached higher accuracy faster than the CNN (see the plots in Figure 10). During training, our method exhibited great variation in the validation plots compared to those of the CNN. This is because the CNN quickly overfitted to the small number of deceased samples but the CNN-AE had to deal with more versatile augmented samples. Thus, the training of the CNN-AE was more difficult, but it achieved better overall performance.

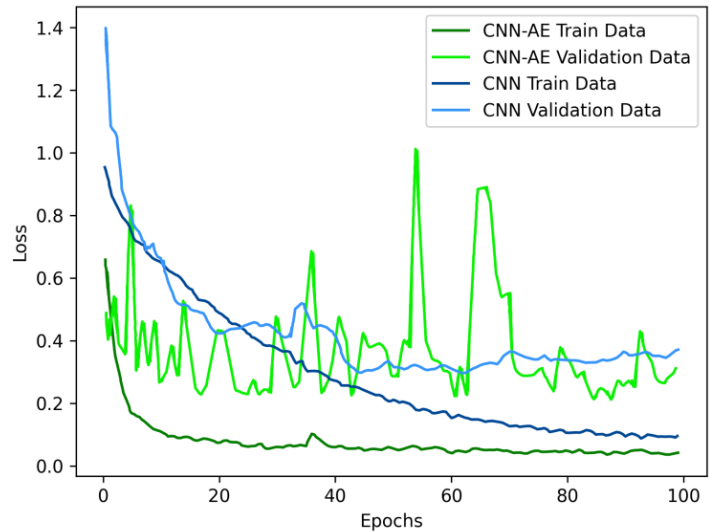

Figure 9. Loss plots of the CNN and CNN-AE methods during the training of our dataset.

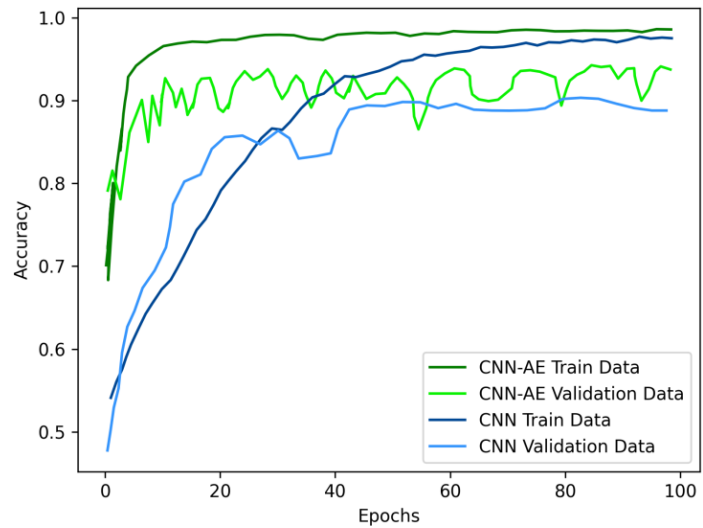

Figure 10. Accuracy plots of the CNN and CNN-AE methods during training of our dataset.

*6.2.2  Comparisons with existing deep models trained on image data*

In this section, we evaluated the performance of various existing deep models that were trained on a dataset of CT images. The CT images were taken from the same patients for whom the clinical dataset was collected. Thus, the results of this section reveal how well deep models trained on CT images perform compared to a CNN trained on clinical data. It should be noted that most of the experiments in the COVID-19 literature revolve around classifying infected and non-infected people using CT images. This section sheds some light on how well deep models can predict the survival chance of already infected patients based on CT images.

The dataset comprised 2822 CT images of recovered patients and 2269 CT images of deceased patients. The CT image dataset size was much greater than the clinical dataset size, as the CT dataset contained multiple images for each patient. As the number of samples of the two classes in the dataset was almost balanced, we did not apply our data augmentation technique to the CT dataset. Additionally, having multiple images for each patient served as a form of data augmentation. This was not the case for the clinical dataset for which each patient had only one value per feature.

In Table 8, the performance metrics for the evaluated deep models are presented as 95% confidence intervals (CIs) that have been computed over a 10-fold cross-validation. The results in Table 8 show that UNet had the best performance among the evaluated methods, followed by Inception Net V3 and DenseNet121, respectively. Overall, Table 8 suggests that some of the famous deep models with pre-trained parameters can be tuned via training to predict the survival chance of COVID-19 patients based on CT images. A performance comparison of the deep models (see Table 8) and the CNN-AE (see Table 7) revealed that a CNN trained on clinical data performed on par with various pre-trained deep models which have been tuned via training on CT data. As stated above, the CT image dataset size was almost 10 times that of the clinical dataset size. However, the CNN trained on clinical data performed almost as well as the deep models trained on CT data. Thus, clinical data could be a good replacement for CT training data if the preparation of the CT images would be difficult or expensive.

Table 8. Results of existing deep models trained on CT images.

| Method | Accuracy (%) | PPV (%) | Recall (%) | Specificity (%) | F1-score (%) | AUC (%) | Loss |
|---|---|---|---|---|---|---|---|
| CNN | 98.88±1.09 | 98.90±1.07 | 98.90±0.91 | 98.10±1.22 | 98.90±0.87 | 98.89±0.92 | 0.01±0.01 |
| DenseNet121 [47] | 99.10±0.10 | 99.00±1.55 | 99.60±0.60 | 98.95±0.36 | 99.20±0.82 | 99.05±1.10 | 0.06±0.07 |
| EfficientNet-B1 [48] | 55.67±1.81 | 56.70±3.32 | 95.40±9.02 | 50.36±2.45 | 70.00±2.34 | 51.16±2.08 | 9.34±5.85 |
| InceptionNet V3 [49] | 99.16±1.26 | 99.80±0.39 | 98.90±1.95 | 99.65±0.22 | 99.40±0.98 | 99.27±1.02 | 0.32±0.61 |
| MobileNet [50] | 75.33±1.62 | 80.70±2.30 | 73.10±1.66 | 79.56±1.48 | 76.60±1.83 | 75.58±1.75 | 0.51±0.02 |
| ResNet50 [51] | 81.63±1.05 | 80.50±1.38 | 88.00±1.01 | 78.36±1.74 | 84.20±1.05 | 80.84±1.06 | 0.46±0.01 |
| VGG19 [52] | 98.02±0.36 | 99.00±0.30 | 97.30±0.72 | 98.79±0.81 | 98.40±0.43 | 98.08±0.35 | 0.07±0.01 |
| Xception [53] | 83.34±0.81 | 93.80±1.20 | 74.90±0.85 | 88.64±0.92 | 83.10±0.94 | 84.36±0.86 | 0.36±0.01 |
| UNet [54] | 99.25±0.21 | 99.80±0.26 | 99.70±0.30 | 98.97±0.19 | 99.70±0.30 | 99.66±0.20 | 0.02±0.01 |

### 6.3 Comparison with other methods trained on clinical data

In this section, we compare the performance of our CNN-AE with some of the existing works on mortality prediction [23, 26, 27]. To this end, we implemented the methods of Chen et al. [23], Zhu et al. [26] and Yan et al. [27]. As mentioned above in the literature review, Chen et al. relied on the RF to assess the severity of COVID-19 patients. For mortality risk prediction, Zhu et al. [26] and Yan et al. [27] used MLP and XGBoost, respectively. These methods were specifically designed to achieve COVID-19-related objectives. For a broader perspective, we also experimented with Naïve Bayes, which is a generic method that can be used regardless of the classification objective. The conducted experiments revealed that our data augmentation approach was generic and beneficial to any classification method.

### 6.3.1 Methods' performance

In this section, we present the experimental results for the classification methods mentioned above. We also investigate the effects of using the proposed data augmentation technique during training. The performance statistics are presented as 95% CIs in Table 9. The CIs are computed based on 10-fold cross-validation. First, each method was trained on the original dataset (without augmentation). The training was repeated using the augmented dataset. The proposed data augmentation using AEs was used for this purpose. All of the rows in Table 9 that are related to training on the augmented dataset are marked with '+AE' postfix in the 'Methods' column. The last row of Table 9 is identical to the last row of Table 7, which has been reproduced here for ease of reference. An inspection of the results in Table 9 reveals that the proposed CNN-AE method outperformed the other methods in terms of accuracy, recall and AUC. Yan et al. [27]+AE, Chen et al. [23]+AE and Zhu et al. [26]+AE claimed second, third and fourth places, respectively. Thus, all methods have clearly benefitted from the augmentation performed on the training dataset. Among the evaluated methods, Naïve Bayes had the worst performance; however, it also benefitted from the augmented dataset.

Table 9. Performance metrics for various classification algorithms with and without AE-based data augmentation.

| Methods | Rank | Accuracy (%) | PPV (%) | Recall (%) | Specificity (%) | F1-score (%) | AUC (%) |
|---|---|---|---|---|---|---|---|
| Chen et al. [23] | 7 | 90.25±3.30 | 93.60±2.67 | 96.30±2.50 | 86.96±2.90 | 94.70±1.83 | 49.82±3.15 |
| Chen et al. [23]+AE | 3 | 95.38±1.40 | 94.50±1.58 | 98.00±1.49 | 92.86±1.65 | 96.10±0.94 | 94.61±2.11 |
| Zhu et al. [26] | 6 | 91.85±1.86 | 94.50±2.25 | 97.50±1.21 | 89.05±1.89 | 95.90±1.03 | 58.67±10.40 |
| Zhu et al. [26]+AE | 4 | 92.97±2.14 | 97.60±1.10 | 90.80±4.39 | 95.06±3.26 | 93.90±2.01 | 93.81±1.73 |
| Yan et al. [27] | 5 | 92.50±2.45 | 94.40±2.32 | 98.00±1.80 | 88.67±2.04 | 95.90±1.29 | 59.17±9.92 |
| Yan et al. [27]+AE | 2 | 95.38±1.28 | 94.50±2.03 | 97.90±2.33 | 91.68±1.85 | 96.00±1.20 | 95.23±1.19 |
| Naïve Bayes | 9 | 61.73±6.93 | 14.30±0.68 | 88.00±5.65 | 42.65±5.95 | 23.90±2.39 | 74.39±5.68 |
| Naïve Bayes+AE | 8 | 74.92±5.49 | 63.80±0.89 | 96.40±3.47 | 60.74±4.45 | 76.50±4.29 | 78.46±4.92 |
| CNN-AE | 1 | 96.05±1.48 | 95.20±1.63 | 98.00±1.33 | 93.13±2.52 | 96.50±1.27 | 95.54±1.70 |

## 6.4 Feature selection analysis

In this section, we examine whether feature selection improves the classification performance of the clinical dataset. We relied on meta-heuristic population-based algorithms to carry out feature selection. The meta-heuristic methods that have been used in the experiments are Artificial Bee Colony (ABC) [55], Ant Colony Optimisation (ACO) [56], Butterfly Optimisation Algorithm (BOA) [57], Elephant Herding Optimisation (EHO) [58], Genetic Algorithm (GA) [59] and Particle Swarm Optimisation (PSO) [60]. Details of the implementation of these methods are available in MEALPY [61], which is a Python module consisting of meta-heuristic algorithms. In all of the experiments detailed in this section, the meta-heuristic methods were run for 500 epochs with a population size of 100.

The results of running each of the meta-heuristic methods listed above was a set of selected features (see Table 10) that specified a subset of the clinical dataset. The dataset extracted subset was used to train a CNN for survival chance prediction. The training was performed with and without data augmentation. The results of the training are presented in Table 11. In each row of the table, the meta-heuristic method used for feature selection and the classifier is specified. Usage of data augmentation is denoted by '–AE'.

Table 10. Selected features by various meta-heuristic methods: (✓) selected feature, (✗) discarded feature.

| Feature Name | ABC | ACO | BOA | EHO | GA | PSO |
|---|---|---|---|---|---|---|
| CBC | ✓ | ✓ | ✓ | ✗ | ✓ | ✗ |
| Blood Type | ✓ | ✓ | ✗ | ✓ | ✓ | ✓ |
| Age | ✓ | ✓ | ✓ | ✓ | ✓ | ✓ |
| Diabetes | ✓ | ✓ | ✗ | ✓ | ✗ | ✓ |
| Blood pressure | ✓ | ✓ | ✗ | ✗ | ✓ | ✓ |
| Asthma | ✓ | ✓ | ✓ | ✗ | ✓ | ✓ |
| Heart disease | ✓ | ✓ | ✓ | ✓ | ✓ | ✓ |
| kidney disease | ✓ | ✓ | ✓ | ✓ | ✓ | ✓ |
| Respiratory disease | ✓ | ✓ | ✗ | ✗ | ✓ | ✗ |
| Cancer | ✓ | ✓ | ✗ | ✗ | ✓ | ✓ |
| Corticosteroids | ✗ | ✓ | ✓ | ✓ | ✓ | ✗ |
| BCG Vaccine | ✓ | ✓ | ✓ | ✓ | ✗ | ✓ |
| Transplant | ✓ | ✗ | ✗ | ✓ | ✓ | ✗ |
| HEM | ✓ | ✓ | ✓ | ✓ | ✓ | ✓ |
| Immunodeficiency | ✓ | ✓ | ✗ | ✗ | ✓ | ✓ |
| Liver disease | ✓ | ✓ | ✓ | ✗ | ✓ | ✓ |
| Rheumatological disease | ✓ | ✓ | ✗ | ✓ | ✓ | ✓ |
| Chest pain | ✓ | ✓ | ✓ | ✓ | ✓ | ✓ |
| Fever | ✓ | ✓ | ✗ | ✗ | ✓ | ✓ |
| Trembling or Shakes | ✓ | ✗ | ✓ | ✗ | ✓ | ✗ |
| Weakness | ✗ | ✓ | ✓ | ✓ | ✗ | ✓ |
| Sweating | ✓ | ✓ | ✓ | ✓ | ✓ | ✓ |
| Sore throat | ✓ | ✓ | ✓ | ✗ | ✓ | ✓ |
| Dyspnea | ✗ | ✓ | ✓ | ✓ | ✓ | ✓ |
| Dry cough | ✓ | ✓ | ✓ | ✓ | ✓ | ✓ |
| Cough with sputum | ✓ | ✓ | ✓ | ✓ | ✓ | ✓ |
| Fatigue, whole body hurts | ✗ | ✗ | ✓ | ✓ | ✓ | ✓ |
| Anosmia | ✓ | ✓ | ✓ | ✗ | ✓ | ✓ |
| Ageusia | ✓ | ✓ | ✗ | ✗ | ✓ | ✓ |
| Anorexia | ✓ | ✓ | ✓ | ✓ | ✓ | ✓ |
| Eczema | ✓ | ✓ | ✓ | ✓ | ✓ | ✓ |
| Conjunctivitis (Pink eye) | ✓ | ✓ | ✓ | ✗ | ✓ | ✓ |
| Blindness and Tunnel vision | ✗ | ✗ | ✗ | ✗ | ✗ | ✗ |
| Vertigo | ✓ | ✓ | ✓ | ✓ | ✓ | ✓ |
| Nausea/Diarrhea | ✓ | ✓ | ✓ | ✓ | ✓ | ✓ |
| Tobacco | ✗ | ✓ | ✓ | ✓ | ✗ | ✓ |
| Gender | ✓ | ✗ | ✓ | ✗ | ✓ | ✓ |

As Table 11 shows, regardless of the feature selection method, the CNN-AE trained on the selected features did not outperform the CNN-AE trained on the full dataset (see the last row of Table 7). This is because the CNN already included an automatic feature selection mechanism and could rule out unnecessary features during learning. Discarding some of the features via feature selection only deprived the CNN of the opportunity to choose the features that best fit its objective.

Among the evaluated feature selection methods in Table 11, BOA showed the best performance, followed by the ACO and ABC, respectively. In relation to Table 11, it should be noted that data augmentation after the application of all of the feature selection methods yielded better results. Thus, the proposed data augmentation approach is generic.

Table 11. CNN and CNN-AE performance trained on features selected by meta-heuristic methods.

| Methods | Rank | Accuracy (%) | PPV (%) | Recall (%) | Specificity (%) | F1-score (%) | AUC (%) | Loss |
|---|---|---|---|---|---|---|---|---|
| ABC+CNN | 8 | 92.32±1.96 | 94.70±1.06 | 97.40±2.23 | 89.65±1.56 | 96.00±1.20 | 53.29±5.79 | 0.25±0.03 |
| ABC+CNN-AE | 3 | 94.61±1.97 | 95.80±2.51 | 95.30±2.37 | 92.92±2.06 | 95.30±1.60 | 94.80±1.95 | 0.24±0.12 |
| ACO+CNN | 7 | 93.10±1.72 | 95.60±1.44 | 97.30±1.92 | 90.57±1.87 | 96.40±0.98 | 62.50±11.63 | 0.26±0.10 |
| ACO+CNN-AE | 2 | 94.71±1.75 | 94.80±1.67 | 96.30±2.50 | 91.95±1.85 | 95.40±1.66 | 94.66±1.83 | 0.22±0.12 |
| BOA+CNN | 12 | 91.37±3.08 | 94.10±2.52 | 97.00±2.48 | 87.08±2.97 | 95.20±1.65 | 53.50±7.11 | 0.28±0.08 |
| BOA+CNN-AE | 1 | 94.99±1.68 | 93.70±2.38 | 97.20±2.46 | 90.82±2.06 | 95.30±1.89 | 94.64±1.81 | 0.24±0.09 |
| EHO+CNN | 10 | 91.86±2.91 | 94.10±2.42 | 98.00±1.49 | 85.69±2.07 | 95.90±1.59 | 53.15±5.63 | 0.23±0.07 |
| EHO+CNN-AE | 5 | 93.95±1.90 | 94.50±2.45 | 95.00±3.98 | 91.57±2.74 | 94.30±1.80 | 93.91±1.82 | 0.26±0.10 |
| GA+CNN | 9 | 92.18±2.08 | 94.80±1.65 | 97.80±1.57 | 88.32±1.76 | 96.10±0.94 | 57.50±10.13 | 0.29±0.07 |
| GA+CNN-AE | 4 | 94.24±1.08 | 94.40±2.17 | 96.20±1.46 | 93.06±1.68 | 95.30±0.83 | 94.47±1.26 | 0.26±0.09 |
| PSO+CNN | 11 | 91.85±2.18 | 95.00±2.71 | 96.40±2.11 | 88.17±1.79 | 95.50±1.25 | 61.51±11.18 | 0.28±0.06 |
| PSO+CNN-AE | 6 | 93.86±2.24 | 94.20±2.87 | 95.20±3.12 | 90.39±2.47 | 94.50±1.94 | 93.10±2.97 | 0.22±0.09 |

# 7 Discussion

This paper focused on survival chance prediction for COVID-19 patients. We performed experiments using both a clinical dataset and a CT image dataset. The size of the CT image dataset was almost 10 times that of the clinical dataset. However, the CNN trained on clinical data performed almost as well as the CNN trained on CT data, which supports the use of clinical data as an alternative for CT images.

Another aspect that might encourage the use of clinical training samples relates to data collection costs. Preparing CT data may require high-end facilities; however, such facilities may increase data collection costs. Additionally, the facilities required to prepare CT data may not be available in deprived areas. Conversely, the tools required to measure clinical data, such as blood pressure, fever and C-reactive protein, are generally accessible.

The proposed method can detect the severity of patients' conditions based on clinical data and enable preventive actions to be taken to minimise the mortality rate. As discussed in Section 2, very few methods have studied mortality rate prediction using clinical data. Additionally, existing methods have used features that differ from the ones we used in our experiments. Thus, the proposed method sheds some light on unexplored aspects of the COVID-19 virus. To implement the proposed system in practice, it must be evaluated by medical experts from medical centres in different regions. After being verified, the system could be used to help experts analyse the severity condition of COVID-19 patients. Thus, patients with critical conditions could be given higher treatment priority than non-critical patients. Prioritising the patients' treatment is of the utmost importance when the medical resources available are limited.

In addition to the proposed method, our dataset can be considered the second contribution of this paper, as it is a good resource for further medical research. The analysis of the importance of the dataset features and their correlations are shown in Figures 3 and 4. Using our dataset, experts can study the relationships between patients' medical conditions (e.g., blood pressure and diabetes) and the likelihood of dying from COVID-19. This will enable medical experts to exercise more caution during the treatment of patients who are more likely to die due to their medical conditions. As the IG values in Figure 3 suggest, there is a strong relationship between the mortality rate of COVID-19 patients and the presence of other critical diseases, such as cancer, kidney and heart diseases. Conversely, mild symptoms and/or diseases, such as dyspnoea, conjunctivitis and asthma, are less likely to contribute to the mortality rate.

Like any other classification approach, the proposed method has some limitations. Due to the use of multiple AEs in the data augmentation phase, the training time of our method was longer than that of a standard CNN. Further, standard CNNs receive a single image sample as input and perform feature extraction automatically. Conversely, we manually collected multiple clinical features for each patient, and such a process is more difficult to manage. Some of the features in our dataset were gathered directly by asking patients; thus, it is possible that patients provided incorrect information.

# 8  Conclusions and future works

In this paper, we investigated the possibility of training a CNN on clinical data to predict the survival chance of COVID-19 patients. To this end, a new dataset consisting of clinical features, such as gender, age, blood pressure and the presence of various diseases, was gathered. The first contribution of this paper relates to our decision to release the collected dataset for public use. We also analysed the dataset features using IG and correlation. Our analysis could aid potential researchers and practitioners with their work on the COVID-19 virus.

To reduce the data imbalance of our dataset, we proposed a novel data augmentation method based on AEs. Our data augmentation approach is generic and applicable to other datasets. Based on the proposed data augmentation approach, a novel survival chance prediction method named CNN-AE was presented, which represents the second contribution of this paper. Using augmented data for training, the 95% CI for the accuracy, recall and specificity of the CNN-AE were $96.05 \pm 1.48\%$, $98.00 \pm 1.33\%$ and $93.13 \pm 2.52\%$, respectively. However, a CNN trained on a dataset without augmentation yielded an accuracy of $92.49 \pm 2.75\%$, a recall of $95.4 \pm 0.88\%$ and a specificity of $96.9 \pm 3.73\%$. Thus, it is clear that the CNN-AE benefitted the data augmentation and outperformed the CNN.

We repeated the CNN training on CT images obtained from the same patients for whom the clinical data had been collected. Comparisons of the performances of the methods trained on clinical data and the methods trained on CT data revealed that clinical data can be used as an alternative to CT images.

In the future, more data needs to be collected to further assess our proposed approach. The use of other data augmentation methods also needs to be investigated and the results compared with our data augmentation method.

**Author contributions**

Contributed to prepare the first draft: R.A., S.H., A.S., F.K., N.H.I., and J.H.J.
Contributed to editing the final draft: S.N., Z.A.S., A.K., S.M.S.I., H.M., and J.M.G.
Contributed to all analysis of the data and produced the results accordingly: D.S., N.H.I., and R.A.
Searched for papers and then extracted data: S.H., A.S., F.K., and J.H.J.
Provided overall guidance and managed the project: S.N., Z.A.S., A.K., S.M.S.I., H.M., and J.M.G.

**Acknowledgement**

This work was partly supported by the Ministerio de Ciencia e Innovación (España)/ FEDER under the RTI2018-098913-B100 project, by the Consejería de Economía, Innovación, Ciencia y Empleo (Junta de Andalucía) and FEDER under CV20-45250 and A-TIC-080-UGR18 projects.